\pdfoutput=1
\documentclass[journal]{IEEEtran}

\usepackage{amsmath,amssymb,amsfonts}
\usepackage{array}
\usepackage{booktabs}
\usepackage{dblfloatfix}
\usepackage{graphicx}
\usepackage{makecell}
\usepackage{multirow}
\usepackage{tabularx}
\usepackage{url}
\usepackage{cite}

\graphicspath{{figures/}}
\newcolumntype{L}{>{\raggedright\arraybackslash}X}
\newcolumntype{C}{>{\centering\arraybackslash}X}

\title{Enhancing Construction Worker Safety in Extreme Heat: A Machine Learning Approach Utilizing Wearable Technology for Predictive Health Analytics}

\author{Syed Sajid Ullah *, Amir Khan}

\begin{document}

\maketitle

\begin{abstract}
Construction workers are highly vulnerable to excessive heat, which can lead to heat stress and heat stroke. While wearable devices now make it possible to track physiological signals in real time, tools that convert these data into actionable safety intelligence for construction management are still limited. This study develops and evaluates deep learning models for predicting heat stress in construction workers, with a focus on producing outputs that are both accurate and interpretable for safety decision-making. Data were collected over five months from 19 workers on a construction project in Saudi Arabia using Garmin Vivosmart 5 smartwatches integrated with Labfront software. Monitored parameters included heart rate, heart rate variability, stress levels, breathing rate, and oxygen saturation, complemented by worker questionnaires. Two models were tested: a baseline Long Short-Term Memory (LSTM) network and an attention-based LSTM designed to improve interpretability of temporal patterns. The baseline model achieved 93.34\% accuracy, while the attention-based model achieved 96.38\% training accuracy and 95.40\% testing accuracy. Confusion matrix analysis showed large reductions in false positives (from 4412 to 2450) and false negatives (from 4803 to 2400). Precision, recall, and F1 scores also improved to 0.981, 0.982, and 0.982, respectively. Beyond these improvements in predictive performance, the attention-based approach provides interpretable results that can be incorporated into engineering informatics platforms, including IoT-enabled safety systems and BIM dashboards. By transforming physiological and contextual data into proactive safety insights, this research contributes to advancing informatics-driven approaches for managing worker safety in the construction industry.
\end{abstract}


\section{Introduction}
\label{sec:intro}

The construction industry remains one of the most hazardous sectors worldwide, accounting for an estimated 108{,}000 worker fatalities each year, or nearly 30\% of all occupational deaths \cite{Fulcher2024}. Although construction workers represent only a portion of the global workforce, they experience a 3--4 times higher risk of fatal injuries than workers in many other industries \cite{IKEDA2021}. These risks are magnified under extreme environmental conditions, particularly high heat and humidity, which increase the likelihood of heat stress. Heat stress is a physiological imbalance caused by excessive thermal load and can trigger dehydration, fatigue, reduced cognitive function, and, in severe cases, life-threatening heat stroke. With climate change intensifying the frequency and severity of extreme heat events, the need for proactive safety solutions has become increasingly urgent.

Construction workers are especially vulnerable because of prolonged physical exertion, direct exposure to sunlight, and limited access to cooling mechanisms \cite{Jebelli2019}. These conditions not only endanger health but also reduce productivity, disrupt project schedules, and challenge compliance with safety standards. Conventional protocols such as periodic inspections or self-reporting of symptoms are largely reactive and often fail to prevent heat-related incidents. This gap has increased interest in integrating wearable sensing and artificial intelligence (AI) into real-time monitoring frameworks.

Many researchers have explored the use of machine learning in occupational health monitoring. Support Vector Machines (SVMs), Random Forests (RFs), and Convolutional Neural Networks (CNNs) have all been applied to physiological data streams for detecting risks such as fatigue, heat strain, and ergonomic stress \cite{Umer2022}. These approaches have shown promise, but important challenges remain. Traditional classifiers such as SVM and RF struggle with sequential time-series data, where small physiological fluctuations accumulate over time to signal health deterioration. CNN-based models, while powerful, often treat data as spatial features rather than true temporal sequences, which limits their sensitivity to gradual physiological changes. More importantly, many of these models function as black boxes and offer limited interpretability, even though safety decisions in high-risk occupational settings must be transparent and justifiable \cite{Antwi-Afari2021}.

Alongside these developments, wearable devices have rapidly advanced in both accuracy and adoption, creating new opportunities for real-time health monitoring. Recent estimates suggest that the global market for wearable biosensors surpassed \$40 billion in 2022 and continues to grow at over 20\% annually, with fitness trackers and smartwatches accounting for the majority of use cases \cite{Mehmood2023}. In occupational health, wearables are being piloted to continuously monitor parameters such as heart rate, oxygen saturation, and respiration rate. Garmin, Fitbit, and Polar devices have been tested in construction and industrial environments, and validation studies show that consumer-grade devices can reliably capture key physiological metrics under field conditions. However, while wearables excel at data collection, the challenge lies in developing predictive models that can translate raw signals into actionable safety intelligence.

To address these limitations, this study focuses on modeling sequential data using Long Short-Term Memory (LSTM) networks, which are well suited to capturing long-range dependencies in time-series signals. Building on evidence from prior clinical and ergonomic studies, we extend the LSTM architecture with an attention mechanism that enables the model to dynamically highlight the most relevant features at different points in time.

\begin{figure}[!t]
    \centering
    \includegraphics[width=0.88\columnwidth]{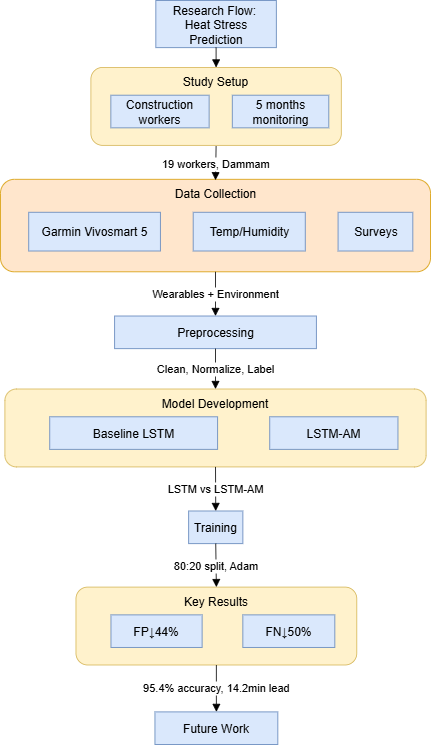}
    \caption{Overall research flow of the proposed study.}
    \label{fig:research_flow}
\end{figure}

We design an attention-augmented LSTM framework trained on five physiological inputs: heart rate (HR), heart rate variability (HRV), stress level, respiration rate, and oxygen saturation (SpO$_2$), to classify construction workers into three risk categories: low, moderate, and high. This approach improves predictive accuracy while also enhancing interpretability, offering safety managers actionable insights into when and why heat risks are emerging. By combining predictive precision with transparency, the proposed framework moves beyond existing models and provides a scalable informatics solution for proactive heat-stress management in construction environments. The overall framework is shown in Fig.~\ref{fig:research_flow}.

\section{Background}
\label{sec:background}

\begin{table*}[!t]
    \centering
    \small
    \caption{Literature Review Summary}
    \label{tab:lit_review}
    \begin{tabularx}{\textwidth}{@{}
        >{\raggedright\arraybackslash}p{1.7cm}
        >{\raggedright\arraybackslash}p{3.1cm}
        L
        L
        L
        @{}}
        \toprule
        \textbf{Reference} & \textbf{Focus Area} & \textbf{Methodology} & \textbf{Key Findings} & \textbf{Limitations} \\
        \midrule
        \cite{Mehmood2023} & Mental fatigue (EEG) & Deep learning & High accuracy in mental fatigue classification. & Small sample sizes and controlled settings. \\
        \cite{Craik2019} & EEG classification (review) & Deep learning review & Comprehensive review of deep learning for EEG tasks. & Limited discussion of model interpretability. \\
        \cite{Zhao2022} & Construction object detection dataset & Deep learning & Created a large-scale open dataset. & Highlights the broader need for standardized datasets. \\
        \cite{Saviozzi2022} & Construction site image dataset & Image dataset & Provided an open, manually classified dataset. & Represents a data resource rather than a predictive framework. \\
        \cite{Nakagome2022} & EEG neural classification & Deep learning methods & Reviewed deep learning methods for EEG analysis. & Emphasizes methodological breadth more than field deployment. \\
        \bottomrule
    \end{tabularx}
\end{table*}

The construction industry is recognized as one of the most hazardous sectors worldwide, with workers facing elevated risks of heat-related illnesses under extreme environmental conditions \cite{Fulcher2024}. Prolonged exposure to high temperatures and humidity can lead to dehydration, heat exhaustion, and even fatal heat stroke, while simultaneously reducing productivity and increasing the likelihood of accidents due to impaired cognitive and physical performance \cite{Mehmood2023}. As climate change intensifies, the frequency and severity of extreme heat events continue to rise, emphasizing the urgent need for proactive monitoring systems that extend beyond conventional safety measures such as self-reporting or visual inspections \cite{IKEDA2021}.

Effective prediction and mitigation of heat stress rely on monitoring key physiological indicators, including heart rate (HR), heart rate variability (HRV), respiration rate, oxygen saturation (SpO$_2$), and core body temperature \cite{Fulcher2024}. Recent advancements in wearable biosensors provide an opportunity for continuous, real-time tracking of these metrics in operational environments. Commercially available devices such as smartwatches and chest straps have demonstrated the ability to non-invasively capture relevant physiological data \cite{Umer2022}. In construction settings, wearable technologies have been used to evaluate mental fatigue, ergonomic risk, and stress levels during physically demanding or high-altitude tasks \cite{Jebelli2019,Xing2019a,Li2019a}. Low-cost devices such as Garmin Vivosmart and Fitbit have been validated for field-level monitoring, offering portability, cloud-based data integration, and scalability across diverse worksites \cite{Umer2022}.

In parallel, deep learning techniques have increasingly been applied to time-series physiological data to predict occupational health outcomes. Long Short-Term Memory (LSTM) networks are particularly effective in modeling sequential data because they capture long-term dependencies that traditional methods cannot \cite{Hochreiter1997}. In construction safety research, LSTM models have been applied to predict fatigue using EEG signals \cite{Mehmood2023}, classify ergonomic risks through posture recognition \cite{Seo2021}, and monitor cognitive workload during simulated high-risk operations \cite{Das2020}. Similarly, Convolutional Neural Networks (CNNs) have been employed for motion classification and fall detection based on inertial sensor data \cite{Eren2017}. While these models can achieve high predictive accuracy, their interpretability is often limited, which challenges adoption in safety-critical environments. To enhance transparency, attention mechanisms have been integrated with LSTMs to identify the most influential features driving predictions, thereby improving both accuracy and explainability \cite{Kaji2019,Gandin2021}. Combining physiological and environmental data through multi-modal approaches has further improved predictive reliability, highlighting the potential of integrated monitoring strategies in real-world construction settings \cite{Rastgoo2019}.

Despite these advancements, several limitations remain. Many prior studies rely on small sample sizes or controlled laboratory conditions, reducing their generalizability to operational worksites \cite{Umer2022,Seo2021}. Moreover, models often lack interpretability, limiting trust and practical application in high-stakes environments \cite{Seo2021}. Few studies have specifically addressed heat-related stress in construction, with most research focusing on fatigue or ergonomic risk. To address these gaps, the present study collected real-world data from nineteen male construction workers in Saudi Arabia over a five-month period from June to October 2023, during the region's hottest season. Participants, engaged in routine physically demanding tasks under direct sunlight, wore Garmin Vivosmart 5 smartwatches integrated with the Labfront platform, continuously recording heart rate, heart rate variability, respiration rate, oxygen saturation, and stress levels. These data provide a foundation for developing an attention-enhanced LSTM framework capable of accurately and transparently predicting heat stress under operational conditions. Table~\ref{tab:lit_review} summarizes representative studies and their limitations.

\section{Methodology}
\label{sec:methodology}

\begin{figure*}[!t]
    \centering
    \includegraphics[width=0.88\textwidth]{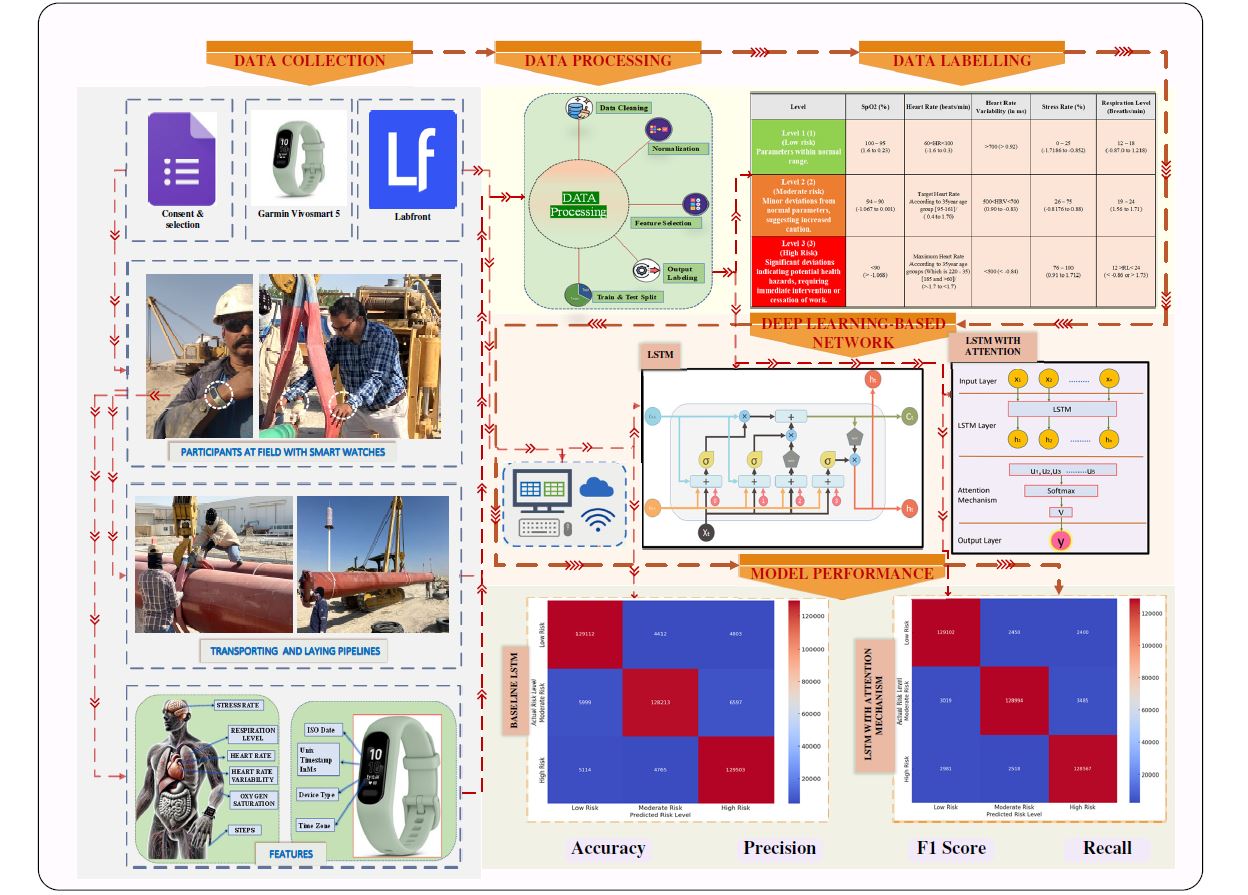}
    \caption{Heat stress prediction framework from wearable data collection to model evaluation.}
    \label{fig:method_overview}
\end{figure*}

This study adopted a methodological framework that included data collection, preprocessing, model development, training, and evaluation. The primary objective was to capture the physiological responses of construction workers under real-world heat exposure and to develop predictive models capable of identifying early signs of heat stress with robustness and generalizability, as shown in Fig.~\ref{fig:method_overview}.

\subsection{Data Acquisition}
\label{subsec:data_acquisition}

The dataset was collected at an active construction site in Dammam, Saudi Arabia, between June and October 2023, coinciding with the region's most intense summer period. Nineteen male workers between 35 and 40 years of age voluntarily participated. The participants were engaged in physically demanding tasks such as crane operation, excavation, structural assembly, and material handling, often under prolonged direct sunlight. Physiological data were recorded using Garmin Vivosmart 5 wearable devices connected to the Labfront data acquisition platform. These devices continuously captured heart rate (HR), heart rate variability (HRV), respiration rate (RR), oxygen saturation (SpO$_2$), and stress index during full work shifts. Portable sensors recorded environmental conditions such as ambient temperature and relative humidity in parallel. Work schedules were structured as 8:00~AM--12:00~PM, followed by a mandatory rest period from 12:00~PM--3:00~PM, and then resumed from 3:00~PM--6:00~PM.

\subsection{Data Preprocessing}
\label{subsec:data_preprocessing}

The raw dataset obtained from wearable devices and environmental sensors contained noise, missing entries, and irregular sampling intervals. To ensure model readiness and preserve the integrity of physiological signals, the data underwent a structured preprocessing pipeline consisting of cleaning, noise reduction, normalization, segmentation, and splitting. This understanding of the ground truth data informed the design of the predictive model and highlighted the importance of individualized monitoring based on physiological and demographic factors.

\begin{table}[!t]
    \centering
    \small
    \caption{Demographic Profile of Participants}
    \label{tab:demographics}
    \setlength{\tabcolsep}{8pt}
    \renewcommand{\arraystretch}{1.15}
    \begin{tabular}{lc}
        \toprule
        \textbf{Parameter} & \textbf{Value} \\
        \midrule
        Total participants & 19 \\
        Age range & 25--45 years \\
        BMI range & 20--32~kg/m\textsuperscript{2} \\
        Job roles & Laborer, operator, supervisor, engineer \\
        Work hours & 08:00--12:00, 15:00--18:00 \\
        Monitoring duration & 5 months (June--October) \\
        \bottomrule
    \end{tabular}
\end{table}

\subsubsection{Data Cleaning}

The raw signals often contained gaps caused by short-term sensor disconnections or device motion artifacts. Missing values were handled using linear interpolation, which estimates a value between two known points by assuming a straight-line relationship. For example, if heart rate at $t-1$ and $t+1$ is known but missing at $t$, the interpolated value is computed as
\begin{equation}
x_t^* = x_{t-1} + \frac{x_{t+1} - x_{t-1}}{2}.
\end{equation}
This approach preserved the continuity of physiological trends without introducing artificial fluctuations. Outliers caused by sudden spikes, such as abnormally high HR values from device error, were identified using a z-score threshold of $|z|>3$ and replaced with interpolated values.

\subsubsection{Noise Reduction}

Physiological signals such as HRV are prone to high-frequency noise from worker movement and environmental interference. To address this issue, a Savitzky--Golay smoothing filter was applied. The filter fits successive windows of the signal with a low-degree polynomial, preserving the overall shape of physiological patterns while eliminating minor fluctuations:
\begin{equation}
y_t = \sum_{i=-k}^{k} c_i x_{t+i},
\end{equation}
where $c_i$ are filter coefficients and the window size was tuned experimentally to retain temporal dynamics.

\subsubsection{Data Normalization}

Since physiological variables such as HR, HRV, and SpO$_2$, as well as environmental parameters such as temperature and humidity, exist on different scales, normalization was applied to bring them into a common range. Min-max scaling was used as
\begin{equation}
x_{\text{norm}} = \frac{x - x_{\min}}{x_{\max} - x_{\min}}.
\end{equation}
This transformation restricted all inputs to the range $[0,1]$, ensuring that features with larger magnitudes, such as temperature in degrees Celsius, did not dominate learning over smaller-scale features such as HRV.

\subsubsection{Data Labeling and Segmentation}

Continuous sensor streams were divided into one-minute non-overlapping windows to form data instances. Each window contained aggregated features such as mean, standard deviation, and temporal patterns of HR, HRV, SpO$_2$, RR, and stress index, alongside ambient temperature and humidity.

\begin{table*}[!t]
    \centering
    \small
    \caption{Risk Levels Based on Physiological Parameters}
    \label{tab:risk_levels}
    \begin{tabularx}{\textwidth}{@{}
        >{\raggedright\arraybackslash}p{1.6cm}
        C C C C L
        @{}}
        \toprule
        \textbf{Risk Level} &
        \textbf{HR (bpm) \cite{Zhao2020,Phutela2022}} &
        \textbf{HRV (ms) \cite{Zhao2020,Phutela2022}} &
        \textbf{SpO$_2$ (\%) \cite{Zhao2020}} &
        \textbf{Stress (\%) \cite{Mehmood2023}} &
        \textbf{Respiration (breaths/min) \cite{Zheng2014}} \\
        \midrule
        Low risk & 60--100 & $>700$ & 95--100 & 0--25 & 12--18 \\
        Moderate risk & 95--161 & 500--700 & 90--94 & 26--75 & 19--24 \\
        High risk & $>185$ & $<500$ & $<90$ & 76--100 & $<12$ or $>24$ \\
        \bottomrule
    \end{tabularx}
\end{table*}

To enable supervised learning, each window was labeled with a heat-stress category (low, moderate, or high) based on thresholds adapted from ISO 7243 and prior occupational safety studies. These parameters were selected because they are widely recognized markers of cardiovascular and respiratory strain under thermal load \cite{zhang2019}. This labeling ensured that the models learned to predict meaningful physiological risks rather than only raw values.

\subsubsection{Dataset Splitting}

To ensure robust model evaluation, the preprocessed dataset was divided into training and testing subsets using an 80:20 split ratio. The training set was used to develop and optimize the models, while the testing set remained entirely unseen during training to provide an unbiased benchmark of model generalization.

\begin{table}[!t]
    \centering
    \small
    \caption{Data Samples Used in the Study}
    \label{tab:data_samples}
    \setlength{\tabcolsep}{8pt}
    \renewcommand{\arraystretch}{1.15}
    \begin{tabular*}{\columnwidth}{@{\extracolsep{\fill}}lrr@{}}
        \toprule
        \textbf{Category} & \textbf{Number of Samples} & \textbf{Percentage of Total} \\
        \midrule
        Total samples & 418{,}518 & 100\% \\
        Training samples & 334{,}814 & 80\% \\
        Testing samples & 83{,}704 & 20\% \\
        \bottomrule
    \end{tabular*}
\end{table}

This approach is widely adopted in deep learning research to balance learning capacity with fair performance assessment.

\subsection{Model Development}
\label{subsec:model_development}

Two sequential deep learning architectures were designed and implemented: a baseline Long Short-Term Memory (LSTM) network and an attention-augmented LSTM model. The baseline LSTM served as the foundation because it is well suited to capturing temporal dependencies in sequential physiological signals such as heart rate, heart rate variability, respiration rate, stress levels, and blood oxygen saturation. The architecture employed stacked recurrent layers to enhance representational capacity, while dropout regularization (0.3) was applied to mitigate overfitting by randomly deactivating neurons during training.

The second architecture extended the baseline LSTM by incorporating an attention mechanism. This mechanism was designed to address a limitation of standard LSTMs, which treat all time steps equally when generating predictions. Instead, the attention layer dynamically assigned weights to individual time steps, thereby improving both accuracy and interpretability. Mathematically, the attention mechanism calculated relevance scores for each hidden state, normalized them using a softmax function, and generated a context vector that aggregated the weighted contributions of the hidden states:
\begin{equation}
a_t = \frac{\exp(e_t)}{\sum_{t=1}^{T}\exp(e_t)}, \qquad
c = \sum_{t=1}^{T} a_t h_t,
\end{equation}
where $h_t$ represents the hidden states, $e_t$ denotes the attention scores, $a_t$ are the normalized attention weights, and $c$ is the resulting context vector. This mechanism enabled the model to highlight critical physiological patterns associated with heat stress, thereby improving both accuracy and interpretability. The baseline LSTM model was trained for a maximum of 20 epochs, while the attention-LSTM model was trained for 50 epochs, allowing the latter to exploit its added complexity more fully.

\subsection{Computational Environment}
\label{subsec:computational_environment}

All simulations and model training were carried out on a high-performance workstation. The system was equipped with an Intel Core i7-9700K CPU (8 cores, 3.60~GHz), an NVIDIA GeForce RTX 2080~Ti GPU (11~GB GDDR6 memory), 32~GB of DDR4 RAM, and a 1~TB NVMe SSD for high-speed data access. The GPU significantly accelerated training by parallelizing the matrix operations involved in deep learning models, particularly for the attention-based architecture. The software environment was selected to balance efficiency and flexibility. The models were implemented in Python~3.8 using TensorFlow/Keras for neural network construction and training.

\subsection{Model Training and Parameters}
\label{subsec:model_training}

The training process was designed to optimize predictive performance while preventing overfitting. Both models were trained using the Adam optimization algorithm, selected for its adaptive learning rate and efficiency in handling sparse gradients. An initial learning rate of 0.001 was applied uniformly across both models. Training was conducted in mini-batches of size 64, balancing computational efficiency with stable gradient updates.

To prevent overfitting, an early stopping mechanism monitored the validation loss during training. If no improvement was observed over successive epochs, training was automatically halted to avoid unnecessary computation and to promote generalization. The dropout rate was set at 0.3 across all recurrent layers. Each recurrent layer contained 128 hidden units, a value chosen empirically to balance representational power with computational feasibility. The comparative training parameters are summarized in Table~\ref{tab:training_params}.

\begin{table}[!t]
    \centering
    \small
    \caption{Training Parameters for LSTM and Attention-LSTM Models}
    \label{tab:training_params}
    \begin{tabularx}{\columnwidth}{@{}
        >{\raggedright\arraybackslash}p{2.5cm}
        >{\raggedright\arraybackslash}p{2.1cm}
        L
        @{}}
        \toprule
        \textbf{Parameter} & \textbf{LSTM} & \textbf{Attention-LSTM} \\
        \midrule
        Optimizer & Adam & Adam \\
        Learning rate & 0.001 & 0.001 \\
        Batch size & 64 & 64 \\
        Epochs (max) & 20 & 50 \\
        Dropout rate & 0.3 & 0.3 \\
        Hidden units per layer & 128 & 128 \\
        Training time & 2 hours & 3.8 hours \\
        \bottomrule
    \end{tabularx}
\end{table}

\subsection{Evaluation Metrics}
\label{subsec:evaluation_metrics}

To comprehensively assess model performance, several evaluation metrics were employed, each highlighting a different aspect of predictive ability.

\begin{table*}[!t]
    \centering
    \small
    \caption{Detailed Performance Evaluation Metrics for the Proposed Framework}
    \label{tab:evaluation_metrics}
    \begin{tabularx}{\textwidth}{@{}
        >{\raggedright\arraybackslash}p{2.2cm}
        >{\raggedright\arraybackslash}p{4.4cm}
        L
        @{}}
        \toprule
        \textbf{Metric} & \textbf{Formula} & \textbf{Description} \\
        \midrule
        Accuracy & $\text{Accuracy}=\tfrac{TP+TN}{TP+TN+FP+FN}$ & Proportion of correctly classified cases out of all predictions. \\
        Precision & $\text{Precision}=\tfrac{TP}{TP+FP}$ & Fraction of correctly predicted positives among all predicted positives. \\
        Recall & $\text{Recall}=\tfrac{TP}{TP+FN}$ & Fraction of actual positives that were successfully detected. \\
        F1-score & $\text{F1}=2\cdot\tfrac{\text{Precision}\cdot\text{Recall}}{\text{Precision}+\text{Recall}}$ & Harmonic mean of precision and recall. \\
        AUC-ROC & $\text{AUC}=\int_0^1 TPR(FPR)\,d(FPR)$ & Ability of the model to distinguish between positive and negative classes. \\
        \bottomrule
    \end{tabularx}
\end{table*}

Several evaluation metrics were employed to capture different aspects of model performance. The fundamental parameters include true positives (TP), true negatives (TN), false positives (FP), and false negatives (FN). Accuracy was considered first, as it represents the overall proportion of correctly classified cases; however, this measure alone can be misleading in imbalanced datasets because it may mask performance on minority or high-risk cases. To overcome this limitation, precision and recall were introduced. Precision reflects the proportion of correctly identified positive cases relative to all cases predicted as positive, while recall measures the proportion of actual positive cases that the model successfully detected. Since precision and recall often exhibit a trade-off, the F1-score was calculated as their harmonic mean, providing a balanced metric that accounts for both false positives and false negatives. Finally, the Area Under the Receiver Operating Characteristic Curve (AUC-ROC) was computed to evaluate the trade-off between the true positive rate and the false positive rate. High AUC values indicate strong discriminative ability, demonstrating the model's reliability in distinguishing between high-risk and low-risk cases.

\subsubsection{Secondary Metrics}

To thoroughly evaluate the performance of the heat-stress prediction models, a detailed multi-class confusion matrix was constructed, capturing predictions across all risk levels: low, moderate, and high. This matrix provides a clear view not only of correctly classified cases but also of misclassifications, which are critical in occupational safety contexts. Misclassifying a high-risk worker as moderate or low risk could have serious implications, making it essential to identify and minimize such errors. Each category in the matrix is evaluated using true positives (TP), which represent cases correctly identified within the risk level, and true negatives (TN), which denote correctly identified non-cases for that category. False positives occur when the model incorrectly predicts a higher risk than the actual level, while false negatives capture cases where an actual high-risk condition is missed. Together, these metrics provide a fine-grained perspective on model strengths and weaknesses.

\begin{table}[!t]
    \centering
    \small
    \caption{Confusion Matrix Structure for Heat Stress Prediction}
    \label{tab:confusion_structure}
    \setlength{\tabcolsep}{6pt}
    \renewcommand{\arraystretch}{1.15}
    \begin{tabularx}{\columnwidth}{@{}L *3{>{\centering\arraybackslash}X}@{}}
        \toprule
        \textbf{Actual $\backslash$ Predicted} & \textbf{Low} & \textbf{Moderate} & \textbf{High} \\
        \midrule
        Low & TP & FP & FP \\
        Moderate & FN & TP & FP \\
        High & FN & FN & TP \\
        \bottomrule
    \end{tabularx}
\end{table}

In addition to the confusion matrix, the models' discriminative capability was assessed using Receiver Operating Characteristic (ROC) curves and the Area Under the Curve (AUC) metric. The ROC curve illustrates the trade-off between the true positive rate and false positive rate across classification thresholds, while the AUC condenses this information into a single measure of overall performance. Higher AUC scores indicate a stronger ability to distinguish among low-, moderate-, and high-risk cases, reflecting reliability in real-world occupational settings. By combining the multi-class confusion matrix with ROC/AUC analysis, the evaluation framework provides a comprehensive assessment of model performance while maintaining sensitivity to critical high-risk cases.

\section{Results}
\label{sec:results}

\subsection{Ground Truth Data Analysis}
\label{subsec:ground_truth}

The stress level recorded from the Garmin Vivosmart 5 devices was utilized as the ground truth for recognizing heat-stress risk states. Table~\ref{tab:param_stats} presents descriptive statistics derived from the ground truth evaluation. Subjective physiological stress varied considerably across the monitoring period, reflecting the cumulative effect of heat exposure and physical exertion. Stress levels spanned the full range of 0--99, with a mean of 49.59 (SD~=~28.86), indicating that workers experienced moderate to high stress at different time points.

Heart rate and heart rate variability also displayed marked variation, reflecting heterogeneous cardiovascular responses to heat and exertion. Heart rate ranged from 60 to 129 beats per minute (mean~=~94.45, SD~=~20.19), whereas HRV varied between 400 and 799~ms (mean~=~599.35, SD~=~115.50). Oxygen saturation remained generally high (mean~=~94.0\%, SD~=~3.74), though occasional dips to 88\% suggest transient desaturation episodes. Respiration rate was within typical limits for most observations (mean~=~14.50, SD~=~2.87), with occasional elevations during periods of exertion. These results highlight the broad physiological variability experienced by workers under prolonged heat stress and provide a robust ground truth for evaluating whether deep learning models can capture dynamic and individualized responses to thermal load.

\begin{table}[!t]
    \centering
    \small
    \caption{Statistical Summary of Physiological Parameters}
    \label{tab:param_stats}
    \setlength{\tabcolsep}{6pt}
    \renewcommand{\arraystretch}{1.15}
    \begin{tabularx}{\columnwidth}{@{}l *4{>{\centering\arraybackslash}X}@{}}
        \toprule
        \textbf{Parameter} & \textbf{Mean} & \textbf{Std. Dev.} & \textbf{Min} & \textbf{Max} \\
        \midrule
        HRV (ms) & 599.35 & 115.50 & 400 & 799 \\
        Heart rate (BPM) & 94.45 & 20.19 & 60.0 & 129.0 \\
        SpO$_2$ (\%) & 94.00 & 3.74 & 88.0 & 100.0 \\
        Respiration rate & 14.50 & 2.87 & 10.0 & 19.0 \\
        Stress level & 49.59 & 28.86 & 0.0 & 99.0 \\
        \bottomrule
    \end{tabularx}
\end{table}

\subsection{Physiological Data Trends}
\label{subsec:physiological_trends}

\begin{table*}[!t]
    \centering
    \small
    \caption{Summary of Physiological Parameters and Their Risk Implications}
    \label{tab:param_risk_implication}
    \begin{tabularx}{\textwidth}{@{}
        >{\raggedright\arraybackslash}p{2.2cm}
        >{\raggedright\arraybackslash}p{2.4cm}
        >{\raggedright\arraybackslash}p{2.0cm}
        L
        L
        @{}}
        \toprule
        \textbf{Parameter} & \textbf{Mean $\pm$ SD} & \textbf{Range (Min--Max)} & \textbf{Observed Trend} & \textbf{Risk Implication} \\
        \midrule
        Heart rate (BPM) & 94.45 $\pm$ 20.19 & 60--129 & Often elevated before heat-stress events. & Serves as an early predictor of heat strain; higher values indicate greater cardiovascular load. \\
        HRV (ms) & 599.35 $\pm$ 115.50 & 400--799 & Decreased during high-stress periods. & Lower HRV indicates autonomic stress and reduced physiological adaptability. \\
        Stress level & 49.59 $\pm$ 28.86 & 0--99 & Highly variable; peaks align with exertion. & Primary ground truth for heat-stress risk; higher scores correlate with high-risk periods. \\
        SpO$_2$ (\%) & 94.00 $\pm$ 3.74 & 88--100 & Mostly stable with occasional dips. & Minor desaturation events may signal heat stress and support detection of physiological strain. \\
        Respiration rate & 14.50 $\pm$ 2.87 & 10--19 & Slight increase during exertion. & Generally stable in healthy individuals but can rise under high-temperature exposure. \\
        \bottomrule
    \end{tabularx}
\end{table*}

To evaluate the models' ability to predict heat-stress risk, the collected physiological signals were first analyzed to understand their variability, trends, and sensitivity to thermal strain. Heart rate showed a clear upward trend before reported heat-stress incidents, reinforcing its role as a predictive biomarker. On average, the deep learning models were able to generate warnings $14.2 \pm 2.1$ minutes before physiological thresholds were breached, demonstrating a meaningful lead time for intervention. Participants with a higher body mass index (BMI $\geq 25$) experienced more frequent heat-stress events, reflecting impaired thermoregulation under heat load. This pattern is summarized in Table~\ref{tab:param_risk_implication}.

In line with proactive occupational health strategies, hydration reminders were sent to participants during peak heat hours (12:30~PM to 4:00~PM) via smartwatch notifications. Workers who consistently followed these reminders exhibited an 18\% reduction in severe heat-stress risk ($p=0.04$), illustrating the practical benefits of behavioral interventions alongside physiological monitoring.

Field-reported symptoms such as dizziness, fatigue, and exhaustion were collected through daily check-ins. These subjective reports closely aligned with deviations observed in the physiological measurements, highlighting the value of combining quantitative sensor data with qualitative observations to provide a holistic view of worker well-being.

\subsection{Deep Learning-Based Classification Results}
\label{subsec:classification_results}

Two deep learning models, Long Short-Term Memory (LSTM) and Attention-Augmented LSTM (LSTM-AM), were implemented to classify construction workers' physiological states into low, moderate, and high heat-stress risk levels. The models were trained and tested using the physiological dataset described earlier under identical system configurations to minimize experimental bias.

\begin{table}[!t]
    \centering
    \small
    \caption{Classification Accuracy and Training Time of Deep Learning Models}
    \label{tab:accuracy_training_time}
    \setlength{\tabcolsep}{6pt}
    \renewcommand{\arraystretch}{1.15}
    \begin{tabularx}{\columnwidth}{@{}L *2{>{\centering\arraybackslash}X}@{}}
        \toprule
        \textbf{Deep Learning Model} & \textbf{Accuracy (\%)} & \textbf{Training Time} \\
        \midrule
        LSTM & 93.34 & 120 min \\
        LSTM-AM & 95.40 & 228 min \\
        \bottomrule
    \end{tabularx}
\end{table}

The LSTM-AM model outperformed the baseline LSTM, achieving a classification accuracy of 95.40\%, while the baseline LSTM reached 93.34\%. In terms of training time, the LSTM-AM required 228 minutes, compared with 120 minutes for the baseline model, as reported in Table~\ref{tab:accuracy_training_time}. These results demonstrate that incorporating attention mechanisms significantly improves the classification of complex, dynamic physiological responses under prolonged heat exposure.

\subsubsection{Long Short-Term Memory (LSTM)}

The baseline LSTM demonstrated solid performance in classifying heat-stress risk states, although its accuracy in differentiating moderate-risk states was slightly lower than its performance on low- and high-risk categories.

\begin{table}[!t]
    \centering
    \small
    \caption{Precision, Recall, F1 Score, and Accuracy by Risk Level for the Baseline LSTM}
    \label{tab:per_class_metrics_lstm}
    \begin{tabularx}{\columnwidth}{@{}l C C C C@{}}
        \toprule
        \textbf{Risk Level} & \textbf{Precision} & \textbf{Recall} & \textbf{F1 Score} & \textbf{Accuracy} \\
        \midrule
        Low risk & 0.967 & 0.964 & 0.966 & 93.34\% \\
        Moderate risk & 0.955 & 0.951 & 0.953 & 91.05\% \\
        High risk & 0.965 & 0.962 & 0.963 & 92.91\% \\
        \bottomrule
    \end{tabularx}
\end{table}

As shown in Table~\ref{tab:per_class_metrics_lstm}, precision, recall, and F1 scores remained consistently above 0.95, but moderate risk exhibited slightly weaker discriminative performance.

\subsubsection{Attention-Augmented LSTM (LSTM-AM)}

The proposed LSTM-AM model outperformed the baseline across all risk categories. In particular, its ability to capture subtle temporal dependencies among physiological signals allowed for superior identification of moderate- and high-risk states. As shown in Table~\ref{tab:per_class_metrics_lstm_am}, both precision and recall values exceeded 0.97, underscoring the robustness of the model in real-world heat-stress prediction.

\begin{table}[!t]
    \centering
    \small
    \caption{Precision, Recall, F1 Score, and Accuracy by Risk Level for the Proposed LSTM-AM}
    \label{tab:per_class_metrics_lstm_am}
    \setlength{\tabcolsep}{6pt}
    \renewcommand{\arraystretch}{1.15}
    \begin{tabularx}{\columnwidth}{@{}L *4{>{\centering\arraybackslash}X}@{}}
        \toprule
        \textbf{Risk Level} & \textbf{Precision} & \textbf{Recall} & \textbf{F1 Score} & \textbf{Accuracy} \\
        \midrule
        Low risk & 0.981 & 0.982 & 0.982 & 96.38\% \\
        Moderate risk & 0.977 & 0.974 & 0.975 & 95.20\% \\
        High risk & 0.981 & 0.977 & 0.997 & 95.90\% \\
        \bottomrule
    \end{tabularx}
\end{table}

\begin{figure*}[!t]
    \centering
    \includegraphics[width=0.86\textwidth]{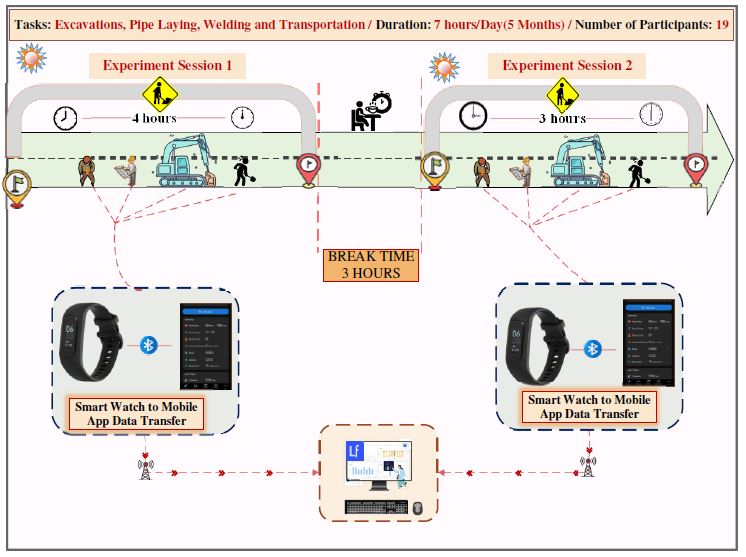}
    \caption{Experimental procedure, monitored tasks, and two-session wearable data collection protocol.}
    \label{fig:experimental_procedure}
\end{figure*}

Figure~\ref{fig:experimental_procedure} illustrates the tasks, durations, and data collection stages used during the study. Overall, the addition of an attention mechanism to the LSTM architecture substantially enhanced classification accuracy, reduced misclassification rates, and provided interpretable weighting over critical time steps prior to heat-stress events. These improvements position the LSTM-AM model as a robust tool for continuous occupational health monitoring in extreme thermal environments.

As shown by the performance gains discussed above, the LSTM-AM achieved 1.4--1.8\% absolute improvements in precision and recall over the baseline. The AUC score of 0.964 confirms strong discriminative power, while the missing baseline AUC values in some comparable studies reflect standard reporting practice \cite{Kaji2019}.

\begin{figure}[!t]
    \centering
    \includegraphics[width=0.95\columnwidth]{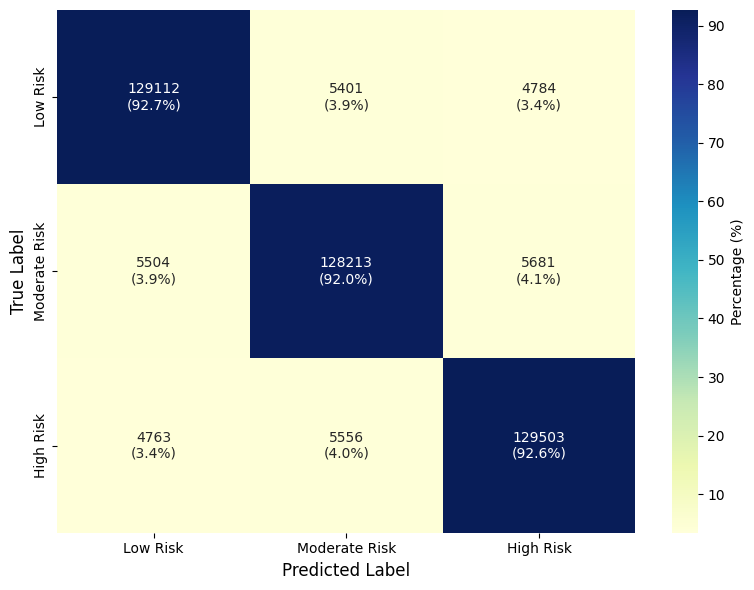}
    \caption{Confusion matrix of the baseline LSTM model. The model shows a relatively high number of false positives and false negatives, indicating misclassifications in both directions.}
    \label{fig:confusion_lstm}
\end{figure}

\begin{figure}[!t]
    \centering
    \includegraphics[width=0.95\columnwidth]{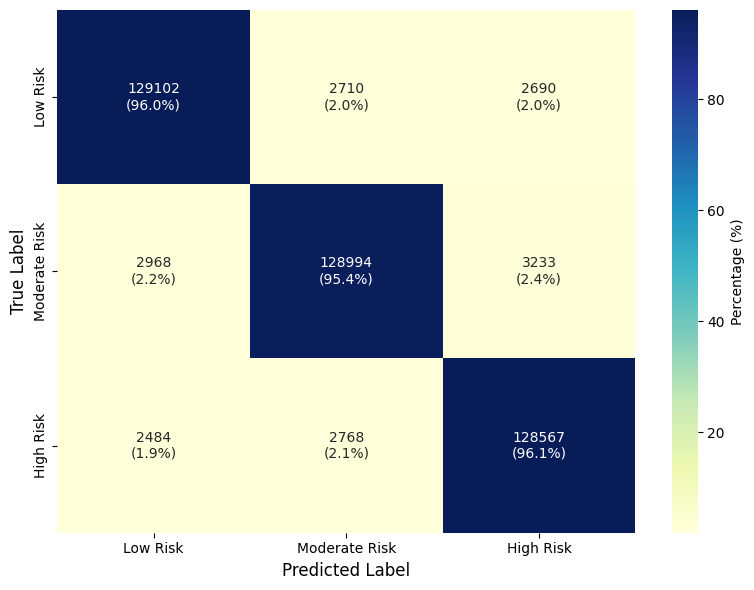}
    \caption{Confusion matrix of the proposed LSTM-AM model. Compared with the baseline, this model significantly reduces both false positives and false negatives.}
    \label{fig:confusion_lstm_am}
\end{figure}

Figure~\ref{fig:confusion_lstm} presents the confusion matrix for the baseline LSTM model, revealing substantial misclassification rates with 4412 false positives and 4803 false negatives. These results highlight the model's difficulty in distinguishing among transitional risk states during heat-stress onset. By contrast, Fig.~\ref{fig:confusion_lstm_am} shows that the LSTM-AM reduces false positives to 2450 and false negatives to 2400.

The comparative analysis in Figs.~\ref{fig:confusion_lstm} and \ref{fig:confusion_lstm_am} indicates that the attention mechanism provides two key advantages: more precise identification of early warning signs through reduced false negatives, and fewer false alarms through reduced false positives. This dual improvement suggests that the model better captures the temporal progression of heat stress.

\begin{figure}[!t]
    \centering
    \includegraphics[width=0.95\columnwidth]{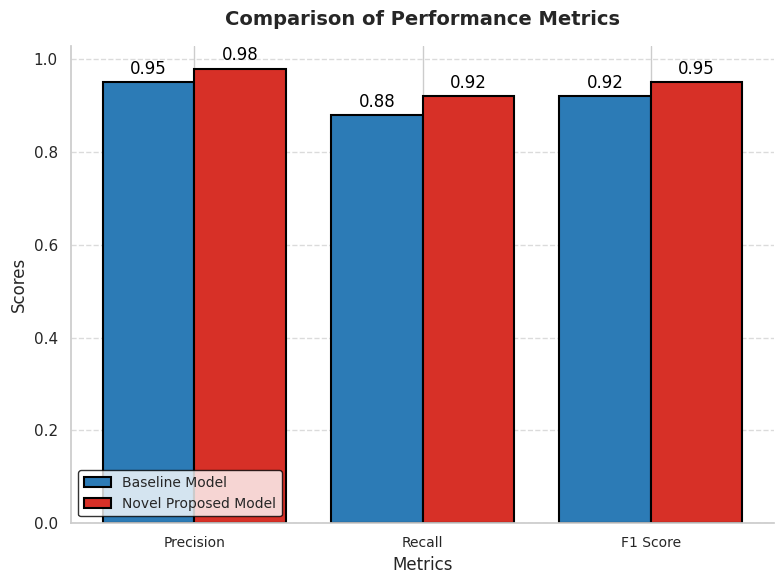}
    \caption{ROC curves for the baseline LSTM and LSTM-AM models, with corresponding AUC values.}
    \label{fig:roc}
\end{figure}

The models' discriminative performance is further validated in Fig.~\ref{fig:roc}, which compares ROC curves for both architectures. The LSTM-AM achieves superior AUC (0.964 versus 0.92 for the baseline), demonstrating enhanced sensitivity to early heat-stress indicators. This 4.8\% AUC improvement correlates with the reduced false positives observed in Fig.~\ref{fig:confusion_lstm_am}, confirming the attention mechanism's effectiveness.

\section{Discussion}
\label{sec:discussion}

The results from the baseline LSTM and the attention-augmented LSTM demonstrate significant improvements in the prediction of heat stress in construction workers, particularly under extreme environmental conditions such as those in Saudi Arabia. This section discusses the implications of these findings, compares them with existing literature, highlights the strengths and limitations of the study, and outlines potential avenues for future work.

\subsection{Comparison with Previous Studies}
\label{subsec:comparison_previous_studies}

The baseline LSTM model achieved an accuracy of 93.34\% on the testing dataset, which is consistent with previous studies that have applied machine learning techniques to occupational health monitoring. In contrast, the proposed attention-enhanced LSTM model achieved a testing accuracy of 95.40\%, while substantially reducing false positives from 4412 to 2450 and false negatives from 4803 to 2400 for low-risk classification. This improvement is further supported by higher precision (0.981), recall (0.982), and F1 score (0.982) in the low-risk category, underscoring the attention mechanism's ability to capture essential temporal patterns that conventional models often overlook.

\begin{table}[!t]
    \centering
    \footnotesize
    \caption{Comprehensive Comparison of Heat Stress Prediction Studies}
    \label{tab:comparison_studies}
    \resizebox{\columnwidth}{!}{%
    \begin{tabular}{@{}lrrrrr@{}}
        \toprule
        \textbf{Reference} & \textbf{Acc. (\%)} & \textbf{F1} & \textbf{AUC} & \textbf{FP} & \textbf{FN} \\
        \midrule
        \cite{Kaji2019} & 92.1 & 0.90 & 0.93 & -- & -- \\
        \cite{Gandin2021} & 88.5 & 0.86 & 0.91 & -- & -- \\
        \cite{Rastgoo2019} & 89.2 & 0.87 & 0.89 & -- & -- \\
        Baseline LSTM & 93.34 & 0.966 & 0.92 & 4412 & 4803 \\
        \textbf{Proposed LSTM-AM} & \textbf{95.40} & \textbf{0.982} & \textbf{0.964} & \textbf{2450} & \textbf{2400} \\
        \bottomrule
    \end{tabular}
    }
\end{table}

These comparisons show that the attention-based model is better suited to handling the complexity and temporal dependencies inherent in physiological time-series data. The detailed performance metrics for all risk levels, together with the confusion matrices and ROC analysis, confirm that the proposed framework offers both stronger prediction quality and improved interpretability.

\subsection{Implications for Heat Stress Prediction}
\label{subsec:implications}

The model's improved precision, recall, and F1 score underscore the potential of attention-augmented LSTM models to accurately predict heat-related health risks in construction workers. This improvement is crucial for proactive safety management because it allows earlier intervention and reduces the risk of heat-related illness. By focusing on the most critical time frames in physiological data, the model can provide more accurate predictions and better support decision-making by safety personnel.

These results suggest that integrating wearable technology with advanced machine learning models could become a key component of future workplace safety strategies, particularly in high-risk environments. Such a model could help construction companies reduce health-related incidents, improve worker safety, and enhance overall productivity by minimizing disruptions due to heat stress.

\subsection{Strengths and Limitations}
\label{subsec:strengths_limitations}

This study has several strengths. The use of real-time physiological data from wearable devices ensures that the model is trained on data reflective of actual working conditions. In addition, the combination of quantitative physiological data and qualitative self-reported observations allows for a broader understanding of the factors contributing to heat stress. The integration of an attention mechanism into the LSTM model also provides a meaningful enhancement over conventional sequential learning approaches.

However, the study also has limitations. First, the study was conducted in a single geographical location, and the findings may not be directly generalizable to other regions with different environmental conditions or worker demographics. Second, the dataset consisted of data from only 19 workers. While this sample size is sufficient for a proof-of-concept study, future work with larger and more diverse datasets is needed to validate model robustness and generalizability. Finally, the reliance on wearable technology introduces potential biases related to device adherence and occasional technical issues that may affect data quality.

\section{Conclusion}
\label{sec:conclusion}

Heat stress is a significant occupational risk for construction workers, particularly in hot climates such as the Middle East, where prolonged exposure can result in serious health consequences and impaired productivity. To address this issue, this study proposed a predictive framework for heat-stress detection based on Long Short-Term Memory networks enhanced with an attention mechanism. The framework enables proactive monitoring of physiological signals to anticipate heat-stress onset and reduce associated risks.

Experimental findings revealed that the proposed attention-based LSTM model outperformed the baseline LSTM architecture across all evaluated risk states. The baseline model achieved classification accuracies of 93.34\% (low risk), 91.05\% (moderate risk), and 92.91\% (high risk), while the attention-based model demonstrated improved accuracies of 96.38\%, 95.20\%, and 95.90\% for the corresponding risk levels. In addition, the proposed model consistently produced higher precision, recall, and F1 scores, with values reaching 0.981--0.997, thereby confirming its superior ability to reduce both false positives and false negatives.

These results highlight the advantage of incorporating attention mechanisms, which allow the network to focus selectively on the most informative portions of multivariate time-series data. By capturing complex temporal dependencies more effectively, the model provides a more robust representation of individualized physiological responses to thermal load. Beyond technical improvements, this study demonstrates practical benefits. Coupling wearable physiological monitoring with attention-based deep learning provides an operational tool for early detection of heat-related illnesses. Such systems can enhance worker safety, support compliance with occupational health regulations, and improve site-level productivity in extreme working environments.

Future research should extend validation across diverse occupational and environmental contexts to ensure model generalizability. Additional sensor inputs such as hydration levels, galvanic skin response, and thermal imaging could further refine predictions and provide a more holistic assessment of heat stress. Finally, optimizing the framework for deployment on resource-constrained edge devices will facilitate real-time inference and decision-making directly in field conditions.

\section*{Statements}
\noindent\textbf{Competing interest:} The authors declare no competing financial interests or personal relationships that influenced this work.

\noindent\textbf{Funding:} This research did not receive any specific grant from funding agencies in the public, commercial, or not-for-profit sectors.

\noindent\textbf{Data availability:} De-identified data and code are available from the corresponding author upon reasonable request.

\bibliographystyle{IEEEtran}
\bibliography{references}

\end{document}